\documentclass[11pt,a4paper]{article}
\usepackage{times,latexsym}
\usepackage{url}
\usepackage[T1]{fontenc}
\usepackage{graphicx}

\usepackage[acceptedwithA]{tacl2021v1}
\usepackage{xspace,mfirstuc,tabulary}

\newif\iftaclinstructions
\taclinstructionsfalse % AUTHORS: do NOT set this to true
\iftaclinstructions
\renewcommand{\confidential}{}
\renewcommand{\anonsubtext}{(No author info supplied here, for consistency with
TACL-submission anonymization requirements)}
\newcommand{\instr}
\fi

\iftaclpubformat % this "if" is set by the choice of options

\else

\fi

\title{LLMs Plagiarize: Ensuring Responsible Sourcing of Large Language Model Training Data Through Knowledge Graph Comparison}

\author{
  Devam Mondal
  \\
    School of Systems and Enterprises, SIT
  \\
  Hoboken, United States
  \\
  \texttt{dmondal@stevens.edu}
  \And
  Carlo Lipizzi 
  \\
    School of Systems and Enterprises, SIT
  \\
  Hoboken, United States
  \\
  \texttt{clipizzi@stevens.edu}
}

\date{}

\begin{document}
\maketitle
\begin{abstract}
  In light of recent legal allegations brought by publishers, newspapers, and other creators of copyrighted corpora against large language model (LLM) developers who use their copyrighted materials for training or fine-tuning purposes, we propose a novel system, a variant of a plagiarism detection system, that assesses whether a knowledge source has been used in the training or fine-tuning of a large language model. Unlike current methods, we utilize an approach that uses Resource Description Framework (RDF) triples to create knowledge graphs from both a source document and a LLM continuation of that document. These graphs are then analyzed with respect to content using cosine similarity and with respect to structure using a normalized version of graph edit distance that shows the degree of isomorphism. Unlike traditional plagiarism systems that focus on content matching and keyword identification between a source and target corpus, our approach enables a broader and more accurate evaluation of similarity  between a source document and LLM continuation by focusing on relationships between ideas and their organization with regards to others. Additionally, our approach does not require access to LLM metrics like perplexity that may be unavailable in closed large language model “black-box” systems, as well as the training corpus. We thus assess whether an LLM has "plagiarized" a corpus in its continuation through similarity measures. A prototype of our system will be found on a hyperlinked GitHub repository.
\end{abstract}
\iftaclpubformat
\fi
\section{Introduction}

Large language models (LLMs) have demonstrated their impressive ability to capture and replicate human language, generating large volumes of text to human-entered prompts. However, this capability exists due to the vast corpora of text these models are trained on. With models exceeding millions of parameters and continuing to grow, they require large volumes of information for training purposes. Sourcing this information, however, presents legal issues due to concerns over the copyright of such corpora. Many argue that copyrighted materials acquired through web-scraping cannot be used in training large language models due to the possibility that such information is regurgitated when producing a result \cite{chang2023speak}. In other words, the LLM plagiarizes this copyrighted content, resulting in legal issues concerning intellectual property and trust. 

Recently, allegations by The New York Times against OpenAI over using its copyrighted content in training ChatGPT brings greater relevance to such an issue \cite{guo2024copyleft}. This paper hopes to tackle this issue by providing a mechanism that assesses whether a piece of content has been used to train/fine-tune a large language model.

Our approach considers the source document (the knowledge base one wishes to see if an LLM has been trained/fine-tuned on), and an LLM continuation of that document when provided the first sentence. We then create RDF (Resource Description Framework) triples in a \texttt{[subject, predicate, object]} format for the source document and the continuation, then make respective knowledge graphs. Then, the cosine similarity of each one-edge walk (a graphical interpretation corresponding to an RDF triple, where the start vertex is the subject, the edge is the predicate, and the end vertex is the object, as demonstrated in Figure 2) of the continuation knowledge graph is compared to each one-edge walk (an RDF triple) of the source knowledge graph. However, since both the source knowledge graph and continuation knowledge graph share their first RDF triple and thus one-edge walk (as they both have the same first sentence), we ignore this triple from the continuation knowledge graph during the comparison. Based on a threshold, if the cumulative cosine similarity is high, there is strong evidence that the source document was used in the training/fine-tuning of the LLM and that the LLM "plagiarized" the source document. We also consider the structure of the two graphs by considering their degree of isomorphism through a normalized graph edit distance metric that considers how many alterations must be made to transform the source knowledge graph into the continuation knowledge graph.

Our method addresses limitations with closed, “black box” large language modeling systems (systems like ChatGPT where metrics of the LLM, as well as training data, are unavailable due to abstraction by the developers) by only considering the outputs of an LLM. Our method also addresses limitations that exist with traditional plagiarism systems (systems that look for similarity between corpora) that utilize direct keyword and content matching by also considering broad relationships between ideas (done so through the aforementioned one-edge walk comparisons) and their presentation/organization with regards to others (done so through assessing the structure of the graph via the degree of isomorphism measured through normalized graph edit distance)

\section{Literature Review}

There exists a large range of literature that addresses identifying large language model training data in a “black-box” environment where the training corpus is unknown. For instance, LLM training data sourcing has been assessed through min-k\% prob, which is a detection method based on the assumption that a member of the training data is less likely to include words that have high negative log-likelihood (and are thus outlier words) compared to a non-member of the training data \cite{shi2024detecting}, therefore considering "anomalous" vocabulary within a text.

	Such an approach, based on the principles of Membership Interference Attacks (MIAs),  an adversarial technique that seeks to determine whether a knowledge source is part of a model’s training data, is the most common method to identify LLM training data in “black-box” environments. Substantial literature also exists about utilizing MIA principles to identify corpora used to fine-tune LLMs, addressing word embeddings \cite{mahloujifar2021membership}, addressing NLP classification models for members of training corpora \cite{shejwalkar2021membership}, and addressing source text memorization \cite{song2019auditing}.

	However, such approaches based on MIA principles take a statistical and probabilistic approach to identifying LLM training data, ignoring other “signs” of sourcing that extend beyond simple copying or paraphrasing. Statistical measures such as only considering the likelihood of "anomalous" words ignore the broad relationships between ideas that exist in sentences of a source corpora that may manifest themselves in an LLM's generated answer. 

	Additionally, traditional plagiarism detection systems (systems that compare the similarity of multiple corpora) often rely on simple matching techniques. For instance, these systems may search direct token (word, sentence, unique phrase, paragraph, etc.) matches between a document and others, using a threshold for matches as an indicator of plagiarism/similarity. Other systems narrow down at the individual word/phrase level, analyzing semantic relationships through simple synonym/antonym detection or more complex Semantic Role Labeling techniques between words in target and source sentences \cite{OSMAN20121493}.  Other systems use character-based n-gram analysis to detect similarity between a target and source corpus \cite{bensalem2014intrinsic}. However, such systems fail to look at similarities in broad idea organization and content structure between a source and target text. They thus do not account for the fact that plagiarism can occur at a high level with regards to ideas and the way they are organized in a text in addition to singular sentence/phrase/word copying. Furthermore, some plagiarism systems utilize Deep Neural Networks in order to assess the similarity between a target and source corpus \cite{ElRashidy2022}, \cite{Hambi2020}. Such systems pose advantages such as the ability to be used with a variety of corpora due to transfer learning concepts such as layer freezing that enables fine-tuning. Furthermore, these systems are able to learn more semantically complex features from the given corpora due to the presence of multiple deep layers. However, due to the black-box nature of such an approach, it is hard to trace what aspects of the target and source match the greatest. 

	Therefore, we hope to augment work in this field by considering knowledge graphs and their ability to model relationships in a transparent way, creating a variant of a plagiarism detection system that can indicate whether a document was used in the training/fine-tuning of an LLM by comparing the similarity between broad ideas present in a source document and an LLM continuation, as well as their organization. Knowledge graphs address the aforementioned limitations with a focus on ideas that extend beyond simple semantic comparison while eliminating a black-box approach to plagiarism detection through easy visualization. 

\section{Approach}
\subsection{Establishing the Ground Truth}
The first part of the system is a source document, the corpus that the LLM is suspected to be trained/fine-tuned on. This is the ground truth for the system and serves as a base for all measurements. To convert this document to a knowledge graph, we first extract RDF triples from the corpus because of their ability to capture complex relationships that encompass the main idea(s) of a sentence. These RDF triples are organized in \texttt{[subject, predicate, object]} format. A knowledge graph $G_S$ can then be produced from a series of RDF triples by looking for a common subject, establishing that as the main starting vertex, and then creating edges from that vertex that correspond to predicates, and then creating additional end vertices that are the objects.

For example, suppose sentence $S$

“A planning process is critical for organizations, assists individuals and benefits society” 

is part of the source document. Using a ChatGPT prompt (found in Appendix A) to extract RDF triples produces the following list:

\texttt{
["planning", "is critical for", "organizations"],
["planning", "assists", "individuals"],
["planning", "benefits", "society"]
}

From here, by establishing the subject \texttt{“planning”} as the central node of the knowledge graph, each edge maps to each unique predicate (\texttt{“is critical for”, “assists”, “benefits”}). Each one-edge walk from the central node thus leads to new end nodes that correspond to each unique object (\texttt{“organizations”, “individuals”,  “society”}) respectively. Figure 1 demonstrates this process.

\begin{figure}
    \centering
    \includegraphics[height=9cm, width=0.55\textwidth]{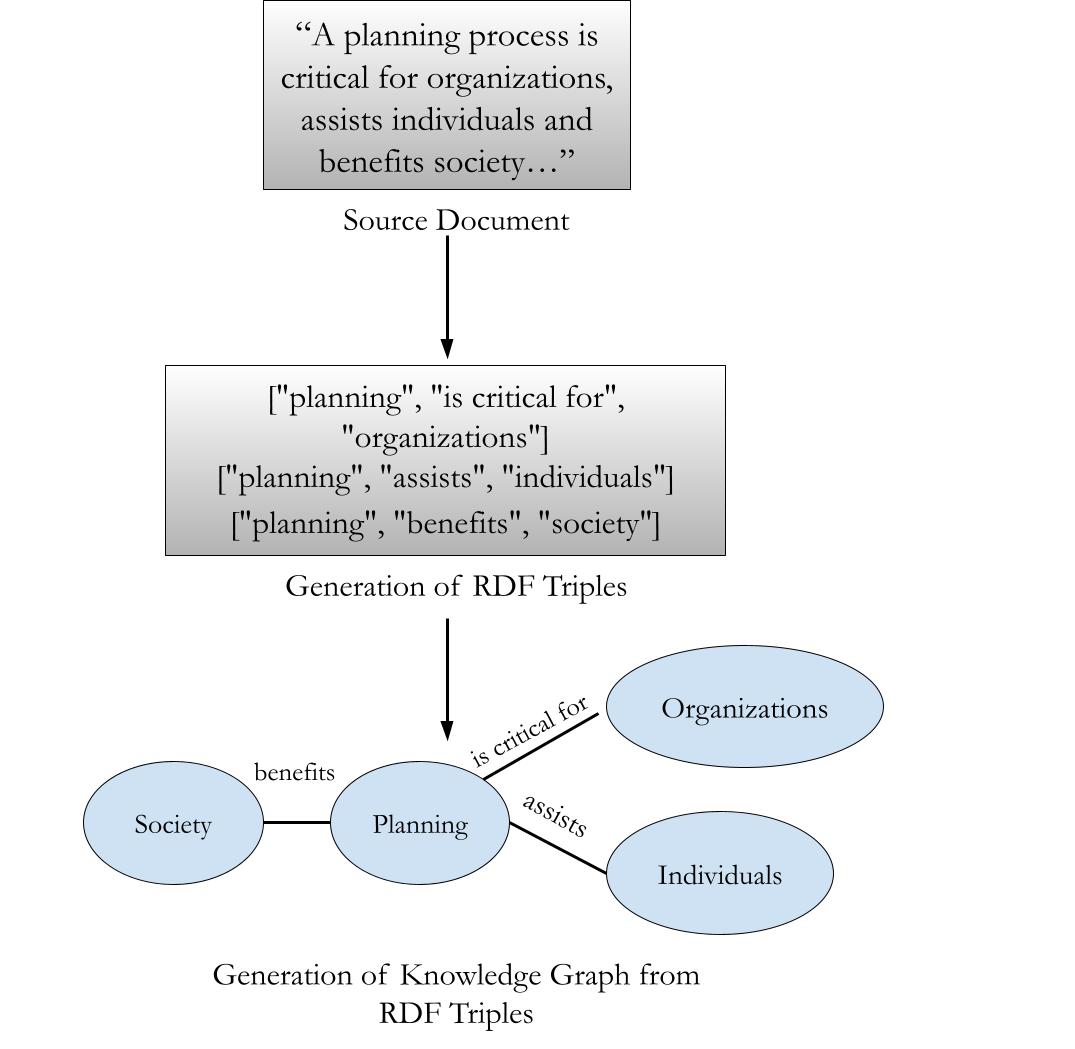}
    \caption{Ground truth knowledge graph generation from a source document using RDF triples.}
    \label{fig:enter-label}
\end{figure}

\subsection{Exposing LLM Understanding of the Ground Truth Using a Continuation}
The second part of this system is obtaining the LLM continuation and generating the continuation knowledge graph. Rather than “quiz” the LLM over content in the source document, we choose to evaluate the continuation to expose the LLM’s thought process with regard to the organization, phrasing, and selection of ideas when given the first sentence of the source document. Furthermore, we believe that generating a continuation is a far more challenging and more insightful test of understanding compared to simple fact retrieval. For instance, if the continuation contains a unique higher-order thought derived through synthesis that is present in the source document, there is a significant chance that the model was trained/fine-tuned on the source document. Simple quizzing would not enable the study of such "emergent properties," ignoring synthesized thoughts LLMs produce through a focus on simple knowledge gathering.

Given source document ${S}$ containing sentences $\{S_0, S_1, S_2…\} \in S$, we provide the LLM the first sentence of the source document ($S_0$) and ask it for a continuation that is $|S| - 1$ sentences long using a ChatGPT prompt (found in Appendix B). The first sentence is then concatenated to the continuation, producing a complete continuation $C$, and the process detailed in Section 3.1 is applied to C to produce knowledge graph $G_C$.

Now, if the LLM continuation has a similar organization, flow of ideas, as well as choice of ideas in an RDF format as the original document, there is a strong likelihood the document was used in training/fine-tuning the LLM. We assess the degree of similarity in flow and choice of ideas through content comparison of the $G_S$ and $G_C$ using cosine similarity. We assess the degree of similarity in organization through the normalized graph edit distance metric that assesses the degree of isomorphism.

\begin{figure*}
    \centering
    \includegraphics[trim={0 4cm 0 6cm},clip,width=\textwidth]{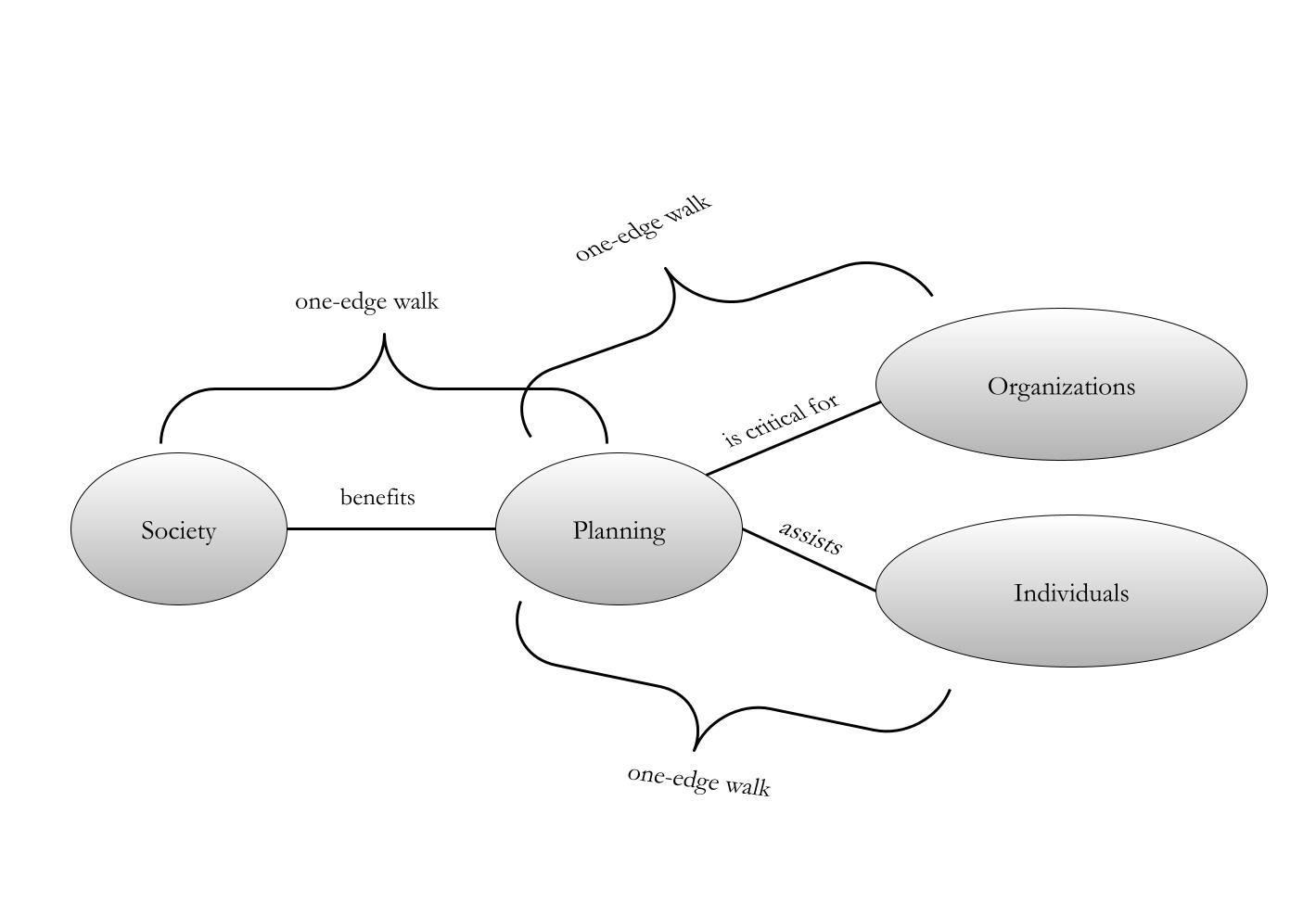}
    \caption{One-edge walks of a knowledge graph.}
    \label{fig:enter-label}
\end{figure*}

\subsection{Assessing Training Usage Through Content}
Once $G_S$ and $G_C$ are created, we consider all one-edge walks of each knowledge graph. Figure 2 shows these one-edge walks in a sample knowledge graph. These walks correspond to the RDF triples created from the source document and the continuation. Both sets of walks, whose vertices ${V}$ are \texttt{\{subject, object\}} and edge $E$ is just \texttt{\{predicate\}}, are then vectorized to embeddings. More specifically, each walk's \texttt{\{subject, predicate, object\}} RDF triple is concatenated and then vectorized. The embeddings for $G_S$ are then stored in a vector database.

From here, we ignore the vectorized walk corresponding to the first sentence of the source document in $G_C$ (as both the source document and continuation share this sentence). We instead take all the other vectorized walks from $G_C$ and find the top three matches with all vectorized walks from $G_S$ with respect to cosine similarity. We choose cosine similarity compared to other similarity metrics (Manhattan, Euclidean, etc.) because it is bounded and is consistently used in other literature when dealing with word embeddings. The similarities for these three matches that are above a user-defined threshold are then added to provide the total similarity $W_{\cos \theta}$ for that vectorized walk. Formally, if $W_{G_C}$ is the vectorized walk from $G_C$, and $m_1$, $m_2$, and $m_3$ are the top three matches from $G_S$ (vectorized walks with the greatest cosine similarity to $W_{G_C}$), under the assumption that all three similarities are over the user-defined threshold, the total similarity ($W_{\cos \theta}$) is:
\begin{equation}
     \frac{W_{G_C} \cdot  m_1}{|W_{G_C}||m_1|} + \frac{W_{G_C} \cdot  m_2}{|W_{G_C}||m_2|} + \frac{W_{G_C} \cdot  m_3}{|W_{G_C}||m_3|}
\end{equation}

Note that there might be situations when an incompatible/nonsensical vectorized walk from $G_C$ only produces two, one, or zero matches. In these situations, Equation 1 reduces to two, one, or zero terms respectively. Additionally, Equation 1 will truncate if there are less than three vectorized one-edge walks in $G_S$, as demonstrated in Section 3.4. 

This process is repeated for all vectorized walks in $G_C$, with every walk’s total similarity adding to the graph’s cumulative similarity $G_{C_{\cos \theta}}$.

From here, we set a threshold for $G_{C_{\cos \theta}}$ based on two factors.

\begin{enumerate}
    \item The minimum required cosine similarity between a $W_{G_C}$ walk and a $W_{G_S}$ walk for it to be considered a match, defined as $min_{\cos \theta}$. This threshold was mentioned earlier in the presentation of Equation 1.  
    \item The minimum total number of vectorized walks from $G_S$ that must produce matches with vectorized walks from $G_C$, defined as $m_t$. This can be calculated by deciding what percentage of all walks need to produce matches, and then multiplying that by the total number of one-edge walks in $G_S$. 
\end{enumerate}

We chose these two factors because they account for sentence-specific similarity (demonstrated by $min_{\cos \theta}$) as well as broader topic-wise similarity (demonstrated by $m_t$) between the source document and the continuation. These two factors are up to the user and are arbitrary values based on the use case.

We therefore define the threshold for similarity $sim_{G_S, G_C}$ as $min_{\cos \theta}$ * $m_t$. If $G_{C_{\cos \theta}} > sim_{G_S, G_C}$, there is strong evidence that the source document has been used to train/fine-tune the model. 

\subsection{Assessing Training Usage Through Content: Example}

To illustrate this component of the system, suppose there exists a source document $S$:

“Supply chains enable individuals to be more efficient. Supply chains fuel business decisions.”

From here, $G_S$ thus consists of the following RDF triples and thus one-edge walks:

\texttt{["supply chains", "enable", "efficient individuals"],
["supply chains", "fuel", "business decisions"]}

Now, consider the following continuation generated by an LLM with the first sentence concatenated at the beginning:

“Supply chains enable individuals to be more efficient. Supply chains optimize decision-making.”

Thus, the RDF triples generated for this continuation that are the one-edge walks for $G_C$ are:

\texttt{["supply chains", "enable", "efficient individuals"],
["supply chains", "optimize", "decision-making"]}

When assessing similarity, we only take the second RDF triple/one-edge walk from $G_C$ (as both $G_S$ and $G_C$ share the first triple). Because there only exist two RDF triples/one-edge walks for $G_S$, there are only two top matches. Equation 1 thus truncates to two terms. With an arbitrary threshold for $min_{\cos \theta}$ set as 0.70 (indicating that a one-edge $G_S$ walk must have a cosine similarity of at least 0.70 with a one-edge walk from $G_C$ for it to be considered a "match"), and $m_t$ as 1 (indicating that the number of total $G_S$ one-edge walk matches must at least be 1), the second RDF triple of $G_C$ generates two matches (with both triples from $G_S$). This is because the cosine similarity between the second $G_C$ one-edge walk and each of the two $G_S$ one-edge walks is greater than 0.70. Equation 1 thus becomes the sums of the cosine similarities of the $G_C$ triple (the second one-edge walk from $G_C$) with the two $G_S$ triples. Note that in this situation, cosine similarity was calculated using \texttt{spaCy} embeddings. For more details regarding the calculations, please refer to Appendix C. In short, the RDF triple for a match is concatenated with appropriate spacing, the RDF triple for a one-edge continuation walk is concatenated with appropriate spacing, and the cosine similarity of the two are calculated:

\begin{equation}
     \frac{W_{G_C} \cdot  m_1}{|W_{G_C}||m_1|} + \frac{W_{G_C} \cdot  m_2}{|W_{G_C}||m_2|}
\end{equation}

\begin{equation}
     0.85 + 0.82 = 1.67
\end{equation}

Since there is only one relevant RDF triple from $G_C$ (the second one-edge walk), its $W_{\cos \theta}$ is the graph's cumulative similarity $G_{C_{\cos \theta}}$. Note that if there were multiple $G_C$ RDF triples (other than the triple that both $G_C$ and $G_S$ share) that produced matches exceeding the $min_{\cos \theta}$ threshold, all of their $W_{\cos \theta}$ values summed would yield $G_{C_{\cos \theta}}$. Therefore, here, the $G_{C_{\cos \theta}}$ is 1.67, greater than the $sim_{G_S, G_C}$ of 0.70 (calculated by multiplying $min_{\cos \theta}$ * $m_t$). Thus, there is strong evidence that suggests the source document was used to fine-tune/train the LLM. Figure 3 provides a visual of this comparison process.

\begin{figure*}
    \centering
    \includegraphics[trim={0 0cm 0 0cm},clip,height=10cm,width=\textwidth]{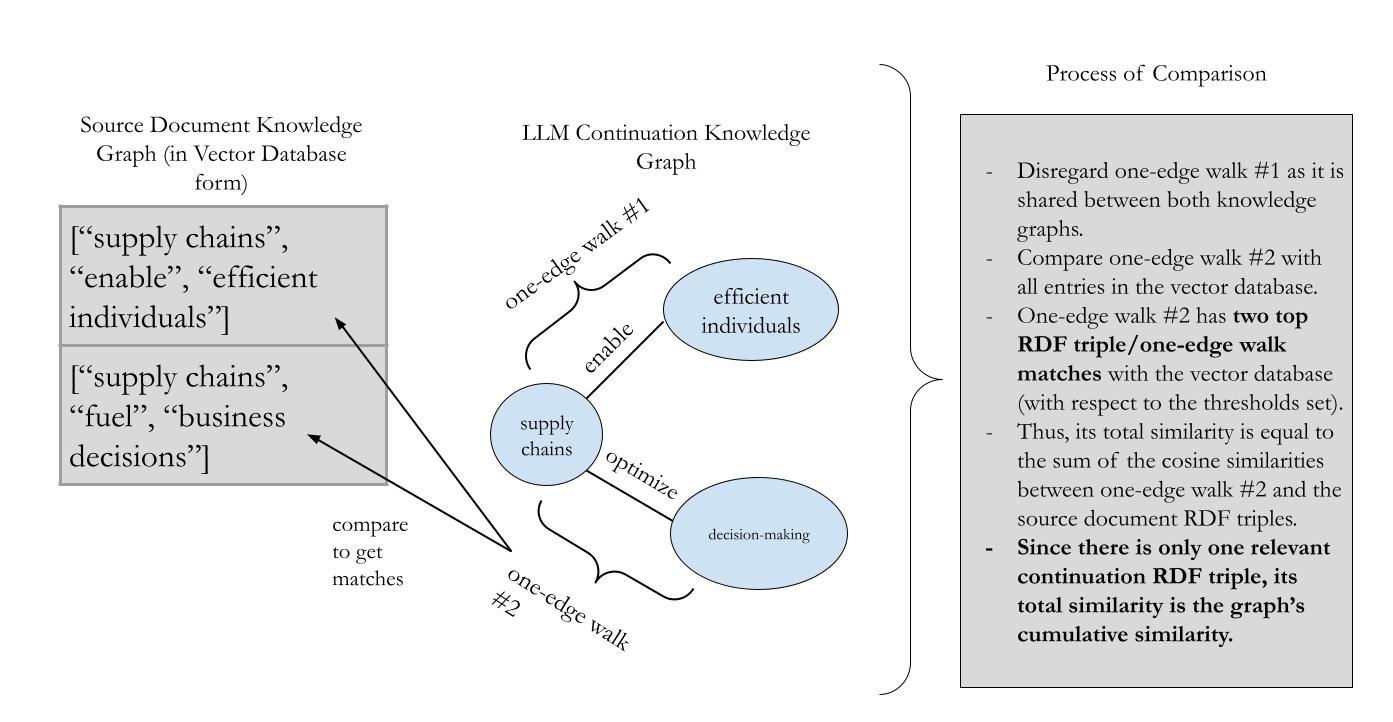}
    \caption{A visual of the comparison process in assessing similarity through content.}
    \label{fig:enter-label}
\end{figure*}

\subsection{Assessing Training Usage Through Content: Nonsensical Example}

To further illustrate this component of the system, we highlight a situation where there is an incompatible vectorized walk from $G_C$. Consider the same source document $S$:

“Supply chains enable individuals to be more efficient. Supply chains fuel business decisions.”

From here, $G_S$ again consists of the following RDF triples and thus one-edge walks:

\texttt{["supply chains", "enable", "efficient individuals"],
["supply chains", "fuel", "business decisions"]}

Now, consider the following continuation, generated by an LLM with the first sentence of the source document concatenated at the beginning:

“Supply chains enable individuals to be more efficient. Supply chains play soccer.”

The RDF triples generated for the LLM continuation that make up the one-edge walks for $G_C$ are the following:

\texttt{["supply chains", "enable", "efficient individuals"],
["supply chains", "play", "soccer"]}

Once again, when assessing similarity, we only take the second RDF triple from $G_C$ (as both $G_S$ and $G_C$ share the first triple). There again exists two RDF triples/one-edge walks for $G_S$,  so there exists the possibility for only two top matches. However, because the second RDF triple from $G_C$ is nonsensical, there exists 0 top matches given the set of thresholds described in Section 3.4. Therefore, the walk’s total similarity $W_{G_C}$ is 0, and the graph’s cumulative similarity is 0. This result makes sense, given the completely nonsensical continuation generated. Figure 4 provides a visual of this comparison process. 

If the second RDF triple was not nonsensical, a situation similar to the one presented in Section 3.4 would occur, and there would be non-zero values for $W_{G_C}$ and $G_{C_{\cos \theta}}$. 

\begin{figure*}
    \centering
    \includegraphics[trim={0 0cm 0 0cm},clip,height=10cm,width=\textwidth]{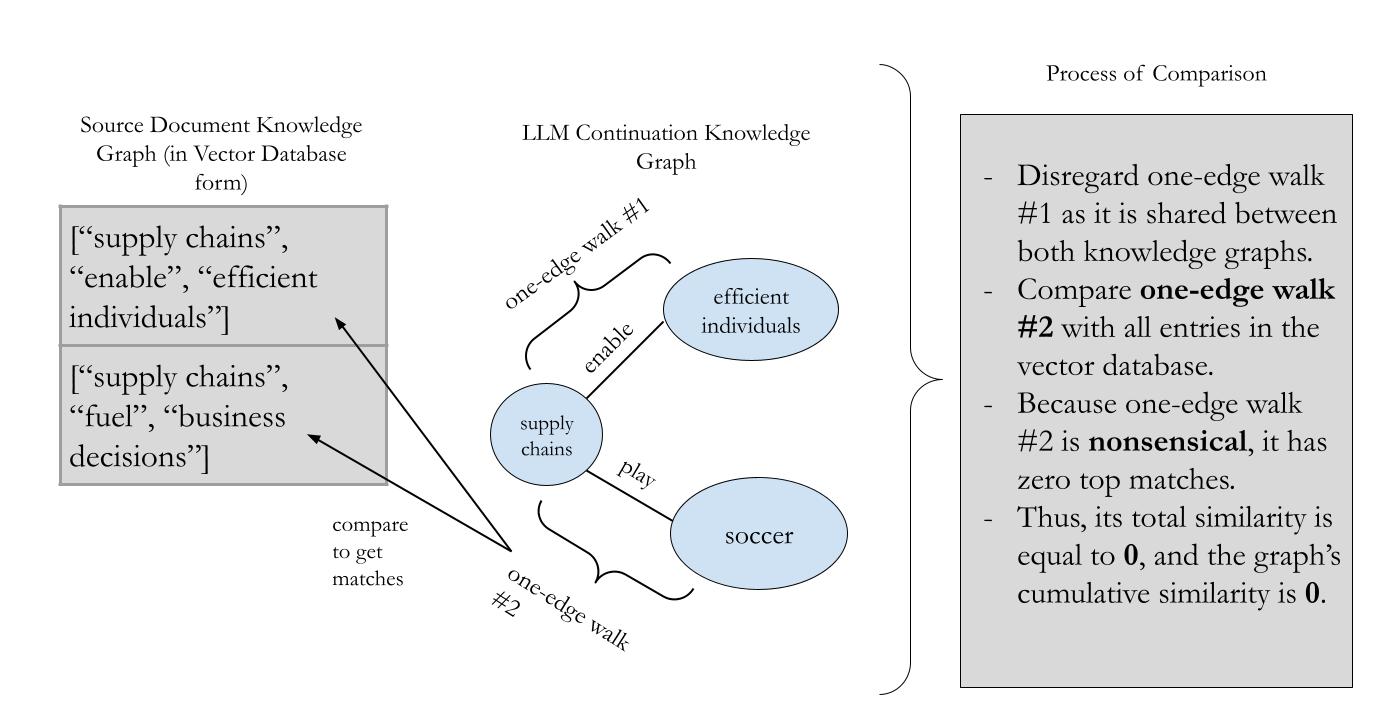}
    \caption{A visual of the comparison process in assessing similarity through content with an incompatible one-edge walk.}
    \label{fig:enter-label}
\end{figure*}

\subsection{Assessing Training Usage Through Structure}

In addition to assessing content similarity between the source document knowledge graph and continuation knowledge graph through cosine similarity, we propose taking into consideration graph structure by considering the two graphs' degree of isomorphism. This is a measure of structural and organizational similarity between the source and the continuation. 

Two graphs can be considered isomorphic if there exists an edge-preserving bijection that enables the one-to-one mapping of two sets of vertices based on equivalent labels. We therefore measure the degree of isomorphism by considering graph edit distance. Graph edit distance is the minimum number of graph edit operations (insertion of nodes, merging/splitting of nodes, edge contraction, etc.) needed to morph one graph into another. As finding graph edit distance is an NP-hard problem, for our purposes, we turn to the most efficient heuristic algorithm when we aim to calculate graph edit distance.

However, because the source document knowledge graph and continuation knowledge graph may not have the same number of edges and vertices in all use cases based on the length of the source document, we propose a relative graph edit distance where the graph edit distance between the source document graph and continuation graph is divided by the sum of the graph edit distance between the source document graph and a null graph ($K_0$) and the graph edit distance between the continuation document graph and a null graph. This is because $n$ number of edits necessary to morph one graph to another when both graphs have a small number of edges and vertices suggests greater structural difference compared to if the two graphs have significantly more edges and vertices. 

Formally, $normGED(G_S, G_C)$ is defined as:
\begin{equation}
   \frac{GED(G_S, G_C)}{GED(G_S, K_0) + GED(G_C, K_0)}
\end{equation}

In addition to $G_{C_{\cos \theta}}$, low values (arbitrary to the user) of $normGED(G_S,G_C)$ must also be taken into account when assessing whether an LLM was trained/fine-tuned on a source document. Lower values of $normGED(G_S,G_C)$ (arbitrary based on the use case) suggest greater structural similarity between the source and continuation knowledge graphs, indicating that the source document was used to train/fine-tune the LLM.

\subsection{Assessing Training Usage Through Structure: Limitations}

It is important to note the major limitation of the $normGED(G_S,G_C)$ metric in that it only considers the structure of two graphs. Thus, this metric cannot alone be used to compare whether $G_S$ and $G_C$ are similar enough to suggest that the source document was used in the training/fine-tuning of an LLM. 

This is because $normGED(G_S,G_C)$ only considers structural similarity and neglects content (the syntactic meaning of the vertices and edges). Its sole usage may thus be misleading. Consider graph $G_F$, made up of three RDF triples/one-edge walks with the central node "stocks," and graph $G_A$ made up of three RDF triples/one-edge walks with the central node "cars." Despite these knowledge graphs being very different when assessed using content, as they focus on two completely different topics, $normGED(G_F,G_A)$ will return 0 because these graphs have the exact same structure (as it requires 0 edit operations to transform one to another). Thus, in the event that a continuation is completely nonsensical yet possesses the same structure and organization of ideas as the source document, analyzing their respective knowledge graphs only through $normGED(G_S,G_C)$ would provide misleading results regarding similarity.

Therefore, it is important to establish some type of weighted compound metric that takes into account both content ($G_{C_{\cos \theta}}$) and structure ($normGED(G_S,G_C)$). 

\section{Summary and Conclusions}
In this paper, we provide a novel system of determining whether a source document was used in the training/fine-tuning of an LLM by considering whether an LLM "plagiarized" the document by analyzing a continuation. Unlike current methods, we leverage knowledge graphs by converting the source document and an LLM continuation of the source document (when given the first sentence) to RDF triples, which are then used to generate a knowledge graph for both. 

After both graphs are vectorized using embeddings, every one-edge walk (which corresponds to an RDF triple) of the continuation knowledge graph, other than the one corresponding to the first sentence (as it is common among both the continuation and source document) is compared to one-edge walks of the source document knowledge graph with respect to cosine similarity. The top three matches (greatest cosine similarities), if above a user-defined threshold, add to a cumulative similarity. This process repeats for all one-edge walks of the continuation knowledge graph, with each total similarity contributing to the graph’s cumulative similarity. When above a user-defined threshold, there is evidence that the source document was used in the fine-tuning/training of the LLM.

In addition to this assessment of content similarity, we propose a new metric to assess structural similarity between the knowledge graph and LLM continuation graph that measures their degree of isomorphism using a relative version of graph edit distance.

In all, our work provides a framework to assess the sourcing of training data for LLMs and helps bring greater accountability for responsible sourcing of training corpora. 

\section{Future Work}

In a follow-up work, we plan to test our system and provide experimental data regarding its effectiveness. We would fine-tune an LLM on a fabricated source document not found on the Internet, then ask the LLM to provide a continuation for that document. We would then compare results on the aforementioned metrics on this fine-tuned LLM with a vanilla LLM.  

Additionally, the thresholds mentioned in our work are up to the user and use case of the document and LLM. We believe finding definite values for various use cases would benefit users of our system and improve decision-making when considering "high" and "low" values for $G_{C_{\cos \theta}}$ and $normGED(G_S,G_C)$.

Furthermore, we have two independent metrics for assessing if a source document was used in the training/fine-tuning of an LLM, one based on content ($G_{C_{\cos \theta}}$) and another based on structure ($normGED(G_S,G_C)$). As mentioned in Section 3.6, it would be beneficial to create a combined metric to produce a single value, where content metrics and degree of isomorphism are both considered. This metric must accordingly weigh the content and structure metric, as two graphs with the same structure may have two completely different meanings and are thus dissimilar. For instance, the continuation and source knowledge graphs may organizationally have similar structures yet have completely different content and meanings. This would suggest that the source document was not used to train/fine-tune the LLM. The metric needs to thus weigh content and structure appropriately.

\bibliography{bibliography}

\begin{thebibliography}{10}
\expandafter\ifx\csname natexlab\endcsname\relax\def\natexlab#1{#1}\fi

\bibitem[{Bensalem et~al.(2014)Bensalem, Rosso, and Chikhi}]{bensalem2014intrinsic}
Imene Bensalem, Paolo Rosso, and Salim Chikhi. 2014.
\newblock Intrinsic plagiarism detection using n-gram classes.
\newblock In \emph{Proceedings of the 2014 Conference on Empirical Methods in Natural Language Processing (EMNLP)}, pages 1459--1464.

\bibitem[{Chang et~al.(2023)Chang, Cramer, Soni, and Bamman}]{chang2023speak}
Kent~K. Chang, Mackenzie Cramer, Sandeep Soni, and David Bamman. 2023.
\newblock \href {http://arxiv.org/abs/2305.00118} {Speak, memory: An archaeology of books known to chatgpt/gpt-4}.

\bibitem[{El-Rashidy et~al.(2022)El-Rashidy, Mohamed, El-Fishawy, and Shouman}]{ElRashidy2022}
Mohamed~A. El-Rashidy, Ramy~G. Mohamed, Nawal~A. El-Fishawy, and Marwa~A. Shouman. 2022.
\newblock \href {https://doi.org/10.1007/s00521-022-07486-w} {Reliable plagiarism detection system based on deep learning approaches}.
\newblock \emph{Neural Computing and Applications}, 34(21):18837–18858.

\bibitem[{Guo et~al.(2024)Guo, Li, Peng, and Wei}]{guo2024copyleft}
Xinwei Guo, Yujun Li, Yafeng Peng, and Xuetao Wei. 2024.
\newblock \href {http://arxiv.org/abs/2402.12216} {Copyleft for alleviating aigc copyright dilemma: What-if analysis, public perception and implications}.

\bibitem[{Hambi and Benabbou(2020)}]{Hambi2020}
El~Mostafa Hambi and Faouzia Benabbou. 2020.
\newblock \href {https://doi.org/10.14569/ijacsa.2020.0110956} {A new online plagiarism detection system based on deep learning}.
\newblock \emph{International Journal of Advanced Computer Science and Applications}, 11(9).

\bibitem[{Mahloujifar et~al.(2021)Mahloujifar, Inan, Chase, Ghosh, and Hasegawa}]{mahloujifar2021membership}
Saeed Mahloujifar, Huseyin~A. Inan, Melissa Chase, Esha Ghosh, and Marcello Hasegawa. 2021.
\newblock \href {http://arxiv.org/abs/2106.11384} {Membership inference on word embedding and beyond}.

\bibitem[{Osman et~al.(2012)Osman, Salim, Binwahlan, Alteeb, and Abuobieda}]{OSMAN20121493}
Ahmed~Hamza Osman, Naomie Salim, Mohammed~Salem Binwahlan, Rihab Alteeb, and Albaraa Abuobieda. 2012.
\newblock \href {https://doi.org/https://doi.org/10.1016/j.asoc.2011.12.021} {An improved plagiarism detection scheme based on semantic role labeling}.
\newblock \emph{Applied Soft Computing}, 12(5):1493--1502.

\bibitem[{Shejwalkar et~al.(2021)Shejwalkar, Inan, Houmansadr, and Sim}]{shejwalkar2021membership}
Virat Shejwalkar, Huseyin~A Inan, Amir Houmansadr, and Robert Sim. 2021.
\newblock \href {https://openreview.net/forum?id=74lwg5oxheC} {Membership inference attacks against {NLP} classification models}.
\newblock In \emph{NeurIPS 2021 Workshop Privacy in Machine Learning}.

\bibitem[{Shi et~al.(2024)Shi, Ajith, Xia, Huang, Liu, Blevins, Chen, and Zettlemoyer}]{shi2024detecting}
Weijia Shi, Anirudh Ajith, Mengzhou Xia, Yangsibo Huang, Daogao Liu, Terra Blevins, Danqi Chen, and Luke Zettlemoyer. 2024.
\newblock \href {http://arxiv.org/abs/2310.16789} {Detecting pretraining data from large language models}.

\bibitem[{Song and Shmatikov(2019)}]{song2019auditing}
Congzheng Song and Vitaly Shmatikov. 2019.
\newblock \href {http://arxiv.org/abs/1811.00513} {Auditing data provenance in text-generation models}.

\end{thebibliography}
\bibliographystyle{acl_natbib}

\appendix
\section{Generating the RDF Triples}

To generate the RDF triples, we utilize OpenAI's ChatGPT 3.5 Turbo API and pass in an "assistant" and "user" message. Below is the base "assistant" message, which serves as the prompt:

\texttt{Given a prompt, extrapolate as many relationships as possible from it and provide a list of updates. provide a json. If an update is a relationship, provide [ENTITY 1, RELATIONSHIP, ENTITY 2]. The relationship is directed, so the order matters. Make the relationship the most granular possible. Examples: prompt: Sun is source of light and heat. It is also source of Vitamin D. updates: [["Sun", "source of", "light"],["Sun", "source of", "heat],["Sun","source of", "Vitamin D"]] prompt: A planning process is critical for organizations, individuals and society.
updates: [["planning", "is critical for", "organizations"],["planning", "is critical for", "individuals"],["planning", "is critical for", "society"]]}

In our prompt, we provided a few examples in order to enable better in-context learning. We then passed in the source document as part of the "user" message. 

\section{Generating the LLM Continuation}

To generate the LLM continuation, we once again used OpenAI's ChatGPT 3.5 Turbo API and passed in the following "user" message:

\texttt{Based off of your training, generate a continuation for the following sentence \textbf{firstLine}. The continuation must EXACTLY be \textbf{sentenceCount-1} sentences long.}

We used a Python \texttt{f} string in order to insert the first line to the "user" message in place of the \texttt{\textbf{firstLine}} variable, as well as to insert the number of sentences the continuation needs to be (inserted in place of the \texttt{\textbf{sentenceCount}} variable).

\section{Other Technical Details}
We used the Python framework NetworkX to create and visualize the knowledge graphs, as well as the OpenAI API to extract RDF triples. We utilized PineCone to create our vector database. 

When calculating the cosine similarity between a match and a continuation one-edge walk, we concatenate the match's \texttt{[subject, predicate, object]} components (separated by spaces), do the same with the continuation one-edge walk, and then calculate the cosine similarity.  

\end{document}